# Identification and Control of a Soft-Robotic Bladder Towards Impedance-Style Haptic Terrain Display

Nathan Baum and Mark A. Minor, *Member, IEEE*

*Abstract*—This paper evaluates the capabilities of a soft robotic pneumatic actuator derived from the terrain display haptic device, "The Smart Shoe." The bladder design of the Smart Shoe is upgraded to include a pressure supply and greater output flow capabilities. A bench top setup is created to rigorously test this new type of actuator. The bandwidth and stiffness capability of this new actuator are evaluated relative to forces and displacements encountered during human gait. Four force vs. displacement profiles relevant to haptic terrain display are proposed and tested using sliding-mode tracking control. It was found that the actuator could sustain a stiffness similar to a soft-soled shoe on concrete, as well as other terrain (sand, dirt, etc.), while the bandwidth of 7.3 Hz fell short of the goal bandwidth of 10 Hz. Compressions of the bladder done at 20 mm/s, which is similar to the speed of human gait, showed promising results in tracking a desired force trajectory. The results in this paper show this actuator is capable of displaying haptic terrain trajectories, providing a basis for futurep wearable haptic terrain display devices.

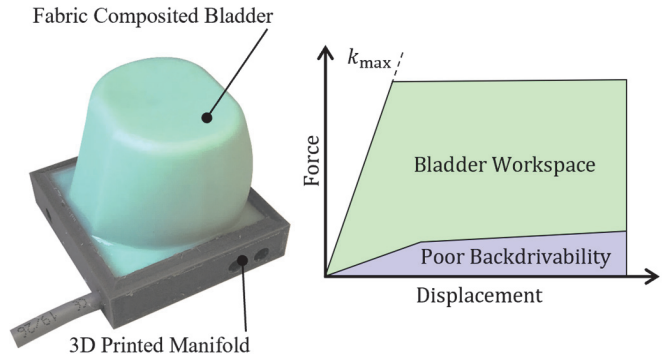

Fig. 1. (Left) A single bladder actuator made of a fabric composited, soft-robotic bladder molded into a 3D printed manifold. (Right) The haptic workspace of a fictitious rubber bladder. $k_{max}$ is the maximum stiffness possible from zero displacement. The poor backdrivability shows the unachievable workspace due to backdrivability issues.

## I. INTRODUCTION

HAPTIC manipulation allows individuals to interact with virtual objects via robotic interfaces [1]. The field of haptics has been thoroughly studied using handheld devices connected to robotic manipulators to interact with virtual objects [2, 3]. Haptic interaction with various terrains through the feet, or *haptic terrain display*, is a developing field. Two systems are currently capable of some form of haptic terrain display. RealWalk [4] used two linear actuators to achieve foot pitch (dorsiflexion and plantar flexion) but was not capable of foot roll (inversion and eversion) due to the limited number of actuators. Re-Step [5] used four linear actuators to apply perturbations creating both foot pitch and roll. However, these two systems cannot render fine terrain features such as a stone under the foot, which would result in displacement at the location of the stone and little to no displacement on the other areas of the foot.

An alternative approach was created using the "Smart Shoe" [6, 7], which consists of an array of soft-robotic, fabric composited [8, 9], elastomeric bladders. The array of five passive bladders allowed pitch and roll of the foot and fine terrain rendering. The rendering capabilities of the Smart Shoe were binary, as in the pneumatic bladder would be opened entirely to simulate no terrain obstacle or would remain closed to simulate an object such as a stone. An updated bladder from the Smart Shoe V4 is used in this work and is shown in Fig. 1.

The work done in this paper is the next logical step in the development of the Smart Shoe. The haptic capabilities of a single bladder actuator are identified by integrating a pressure supply into the design and using pulse-width modulation (PWM) closed-loop control.

A basic illustration of the actuator force vs. displacement haptic workspace is shown in Fig. 1. The green section shows the achievable forces and displacements from the actuator. The blue section shows a portion of the workspace that is unachievable due to the mechanical stiffness of the bladder walls. This work determines this haptic workspace experimentally while also testing the bladder on various force vs. displacement curves in the actuator workspace that are relevant to haptic terrain display.

In this work, we find two metrics to be most important. First, there exists some stiffness, $k_{req}$, which is the required actuator stiffness for successful terrain display. If a user of the Smart Shoe is walking over concrete, each actuator should be capable of displaying the forces from the concrete as well as the virtual pair of shoes worn by the user. For stiff surfaces such as concrete, the equivalent stiffness of the two surfaces in series is approximately the stiffness of the shoe sole. Less stiff terrain would result in a lesser stiffness than the shoe sole. Therefore, the required stiffness for successful terrain display, $k_{req}$, is approximately the stiffness of a shoe sole. In this work, we determine this value for a foam shoe sole.

The second metric is the actuator bandwidth relative to the frequency spectrum of ground reaction forces in gait. Ground reaction forces (GRF) have been studied in [10] and modeled in [11], which found a majority of vertical GRF (vGRF) were contained under 10 Hz, with the heel strike transient (HST) occurring at nearly 75 Hz. The success of the actuator bandwidth should be measured relative to these values.

This work is the first to evaluate a soft-robotic bladder actuator intended for haptic terrain display relative to the

This work was supported by NSF grant No. 1162617.
N. Baum and M. Minor are with the Department of Mechanical Engineering, University of Utah, Salt Lake City, Utah 84112, USA; Emails: Nathan Baum: nate.baum@utah.edu, Mark Minor: mark.minor@utah.edu.

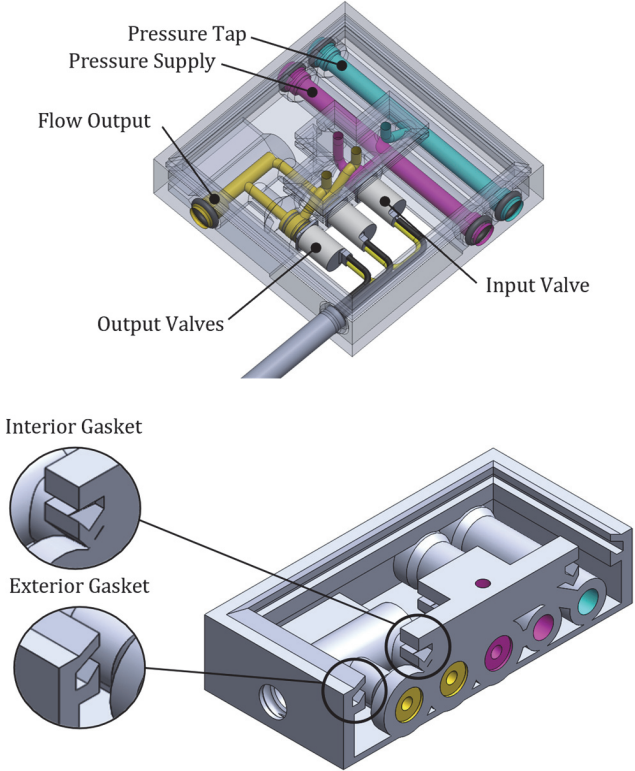

Fig. 2. (Top) The internal flow paths of the bladder actuator. A pressure tap line (cyan) allows measurement of internal bladder pressure. A pressure supply line (magenta) allows access to the 10 psi pressure supply valve via the input solenoid valve. Flow output (yellow) lets air exit the bladder through the two output solenoid valves. (Bottom) Interior and exterior grooves are added to the design for sealing.

metrics of required actuator stiffness and bandwidth during gait. Additionally, this work proposes a new force model that improves upon previous work in modeling this style of soft-robotic bladder and employs a sliding-mode tracking controller used in an impedance-style control scheme.

The remainder of this paper begins by describing the bladder actuator and testing setup, which is followed by the identification of the force model, PWM valve flow, and closed-loop bandwidth. The sliding-mode tracking controller is then designed and various curves in the bladder workspace are tested at various compression rates. The result of this testing is presented in the following section. The paper concludes with a discussion of these results.

## II. Bladder Construction

The bladder is constructed from fabric composited silicone rubber walls molded into a 3D printed manifold. The actuator manifold interior is shown in Fig. 2. Three solenoid valves are used to distribute air into and out of the bladder. Two solenoid valves control the output flow (yellow) while the third inputs air into the bladder from the 10 psi pressure supply (magenta). A pressure tap line allows for external pressure sensing (cyan). Interior and exterior grooves fill with silicone rubber during the molding process to act as sealing gaskets.

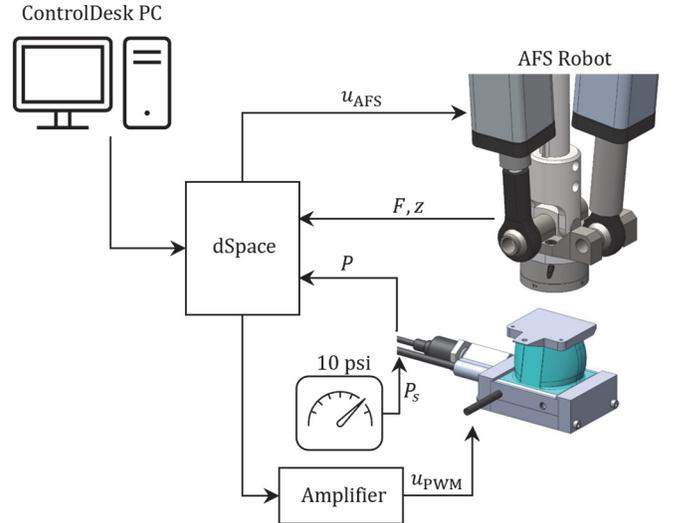

Fig. 3. The testing setup for the bladder actuators. ControlDesk is run on a PC and interfaced with a dSpace MicroLabBox. The AFS is controlled by $u_{\text{AFS}}$, while the bladders are controlled by $u_{\text{PWM}}$. An amplifier of custom design circuitry drives the solenoid valves inside the actuator. A pressure supply, $P_s$, supplies 10 psi to the bladder. A pressure sensor returns pressure measurement, $P$, to the dSpace. The AFS robot measures bladder compression, $z$, and bladder force, $F$, which is sent back to the dSpace system.

## III. Testbed Setup

The testing apparatus used for the system identification and control is illustrated in Fig. 3. The bladder actuator is controlled from a dSpace system by control input, $u_{\text{PWM}}$, which is a voltage sent to the interior solenoid valves. A pressure supply, $P_s$, provides pressure of 10 psi to the bladder while a pressure sensors returns internal pressure, $P$, to the bladder. The Ankle-Foot Simulator (AFS) [12, 13] robot is controlled by the dSpace system by control input, $u_{\text{AFS}}$. The AFS robot returns measurements of bladder compression, $z$, and bladder force, $F$. The dSpace system is interfaced with a PC running ControlDesk and Simulink.

## IV. System Identification

This work identifies all the models necessary for model-based sliding-mode control of the soft-robotic bladders including bladder force, input and output flow, and closed-loop bandwidth.

### A. Force Modeling

A free-body diagram of the bladder is shown in Fig. 4. A bladder actuator at some compression, $z$, and internal pressure, $P$, is subjected to five forces: the stiffness of the air inside the bladder, $k_a$, applies air pressure force, $F_a(z, P)$; the stiffness of the fabric-composited silicone walls result in the wall force, $F_{kw}(z, P)$; various damping terms associated with the walls and air pressure are lumped into $b_{eq}$, which gives damping force, $F_b(z, \dot{z}, P)$, a function of compression velocity, $\dot{z}$; the gravity vector, $\boldsymbol{g}$, applies a gravity force, $F_g$, to the bladder; the external force applied to the bladder is written as $F(t)$.

The system states can be written as

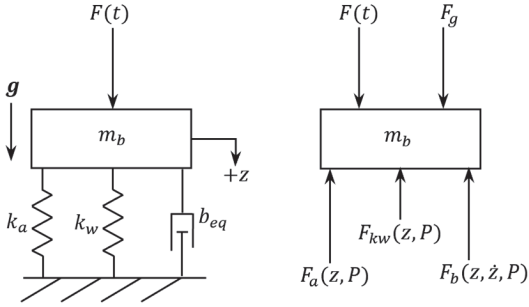

Fig. 4. The free-body diagram the describes the bladder forces.

$$\dot{z} = v$$
$$\dot{v} = \frac{1}{m_b}\left[F(t) + F_g - F_{kw} - F_b - F_a\right] \quad (1)$$
$$\dot{P} = \frac{P}{V}(Q - \dot{V}),$$

where $v$ is compression velocity. The pressure dynamics, $\dot{P}$, include control input of flow, $Q$, bladder volume, $V$, and the change in volume, $\dot{V}$. This equation is derived from the ideal gas law assuming adiabatic conditions.

Previous work [6] used a complex lumped parameter model based on a mass-spring-damper with many parameters solved for using a root-mean-square error (RMSE) fit to estimate $F_{kw}$ and $F_a$. The bladder walls were simulated in small segments using finite element analysis (FEA) at each timestep, making this model an unlikely candidate for a real-time control application.

Alternatively, this work proposes a polynomial fit to collected force, pressure, and displacement data, which was found to be more accurate, more applicable to a real-time application, and easier to use in model-based control schemes. The actuator force, $F(t)$, can be written as

$$F(t) = f(z, \dot{z}, \ddot{z}, P, t) \quad (2)$$

where $z$ is bladder displacement, time derivatives $\dot{z}$ and $\ddot{z}$ are velocity and acceleration, respectively, $P$ is the internal bladder pressure, and $t$ is time. The updated system states are written as

$$\dot{z} = v$$
$$\dot{v} = \frac{1}{m_b} f(z, \dot{z}, \ddot{z}, P, t) \quad (3)$$
$$\dot{P} = \frac{P}{V}(Q - \dot{V}).$$

Data was collected using the testbed setup shown in Fig. 3. The AFS robot performed a quasi-static compression of the bladder for 20 mm. This compression test was repeated for bladder pressures from 0 psi to 10 psi in 1 psi increments.

A polynomial model was fit to the collected force data. The polynomial fit is shown in Fig. 5, with the experimental data shown. The polynomial fit resulted in an $R^2$ of 0.988. The polynomial is third order with respect to $z$ and first order with respect to $P$ as well as constrained by $\hat{F}(z=0, P=0) = 0$. The final model is written as

$$\hat{F} = 21.52z - 2.245z^2 + 3.028zP + 0.07648z^3 \\ - 0.1165z^2 P. \quad (4)$$

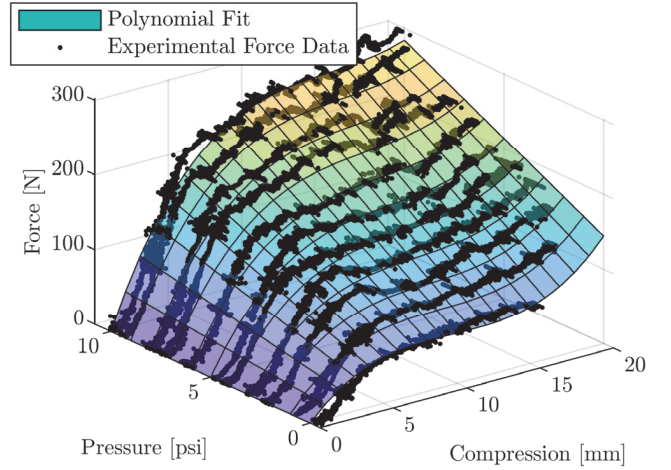

Fig. 5. The model fit for the function $\hat{F} = \hat{f}(z, P)$ where $\hat{F}$ is an estimate of bladder force, $z$ is bladder compression, and $P$ is bladder internal pressure. The model fit is a third-order polynomial with respect to $z$ and first order with respect to $P$.

### B. Valve Flow Modeling

The manufacturer specifies valve flow, $Q$, as

$$Q = \frac{2K f_T}{Lohms}\sqrt{(P_a - P_b)P_b} \quad \begin{array}{l} P_a/P_b < 1.9 \\ \text{(subsonic region)} \end{array} \quad (5)$$

where $K$ and $f_T$ are supplied constants, $P_a$ and $P_b$ are upstream and downstream pressures, respectively, and $Lohms$ is a flow resistance value from [14]. For this system, pressures always operate in the subsonic region. PWM is used to relate $Lohms$ to PWM duty cycle, which is a common method used in pneumatic actuators [15, 16, 17, 18, 19, 20] to change the flow resistance in the valve to provide control over the flow.

The output port of a single actuator is connected to a flow meter (Omron D6F-20A6) to measure output flow. The actuator is connected to a pressure supply set at 10 psi, where the PWM duty cycle is increased from 0% to 100% before decreasing in the opposite direction over 20 seconds. This procedure is repeated in 1 psi increments down to 1 psi. Multiple PWM frequencies are tested between 10 – 50 Hz.

The collected data is processed by doing an RMSE search for the three coefficients that best fit a power model and minimize RMSE. The best results were found using a PWM pulse frequency between 25 – 40 Hz. The output flow model that computes the PWM duty cycle ($DC$) as a function of desired flow, $Q_{des}$, atmospheric pressure, $P_{atm}$, the internal pressure, $P$, is written as

$$DC_{out} = 7245 * \left(\frac{2K f_T \sqrt{(P - P_{atm})P_{atm}}}{Q_{des}}\right)^{-0.7814} + 12.6. \quad (6)$$

This power model is shown graphically in Fig. 6.

The actuator is then connected for input flow testing, where the flow meter is placed in line with the 10 psi pressure supply. The actuator is set to atmospheric pressure before the air can enter from the pressure supply line. This procedure is repeated

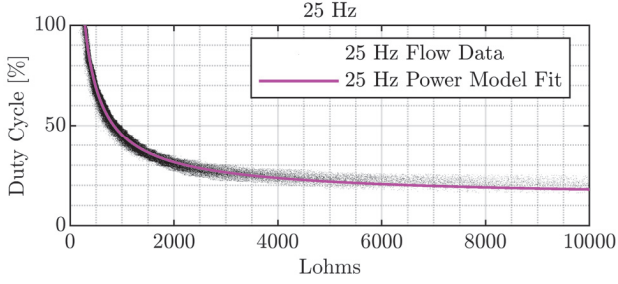

Fig. 6. The power model that relates output flow resistance, $Lohms$, to PWM duty cycle.

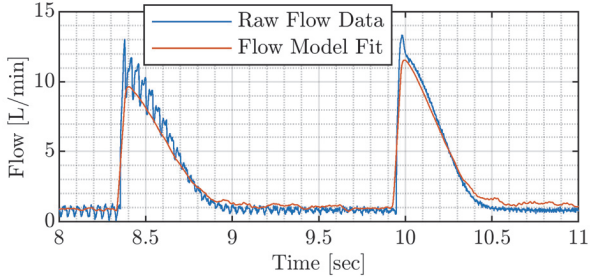

Fig. 7. The data fit for input flow into the bladder actuator.

for PWM frequencies between 10 – 50 Hz and duty cycles between 10% and 100% in 10% increments.

The duty cycle for the desired input flow is written as

$$DC_{in} = 6000 * \left(\frac{2Kf_T\sqrt{(P_s - P)P}}{Q_{des}}\right)^{-0.7} + 16. \quad (7)$$

where $P_s$ is the 10 psi pressure supply. The curve fit to the raw data is shown in Fig. 7.

### C. Closed-Loop Bandwidth

A common technique of determining system bandwidth is by performing a *frequency sweep*. A frequency sweep is a technique where a sinusoidal input is applied to the system, and the output is recorded. The sinusoid frequency is increased until the output magnitude has broken down. The magnitude and phase shift of the output relative to the input is recorded and plotted as a Bode plot.

A pressure controller was tuned for the best possible performance to a step input. A desired pressure sinusoidal trajectory was input into the controller, and the output was recorded. This experiment was conducted for an uncompressed bladder with a pressure supply of 10 psi at three PWM frequencies: 10 Hz, 25 Hz, and 40 Hz.

The results of the frequency sweep are shown in Fig. 8. The experimental data was fit using nonlinear least squares to a second-order transfer function

$$\hat{G}(s) = \frac{\hat{\omega}_n^2}{s^2 + 2\hat{\zeta}\hat{\omega}_n s + \hat{\omega}_n^2} \quad (8)$$

where $\hat{\omega}_n$ is an estimate of the natural frequency, $\hat{\zeta}$ is an estimate of the damping factor, and $s = j\omega$, which is an input of frequency $\omega$. The bandwidth of the estimated transfer function is computed by finding where the magnitude crosses the -3 dB value. The highest bandwidth was achieved with a PWM pulse frequency of 40 Hz. The estimated second order parameters were computed as $\hat{\zeta} = 0.86$ and $\hat{\omega}_n = 58.5$ rad/s with a bandwidth of $\omega_{BW,40} = 7.3$ Hz.

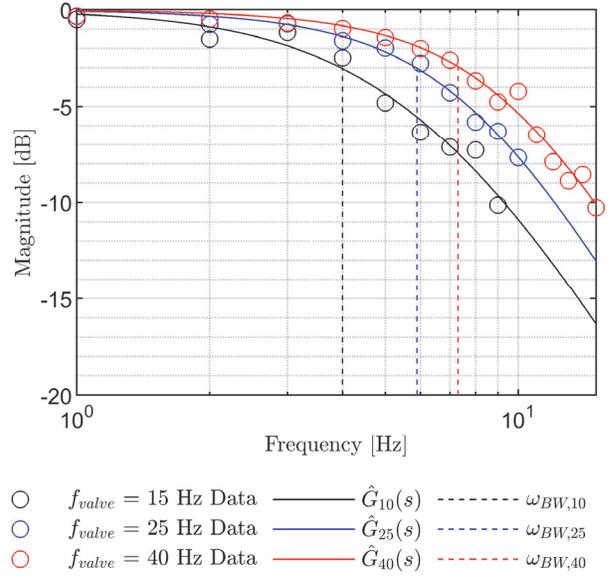

Fig. 8. The magnitude portion of Bode plots for the closed-loop system with a desired pressure input. Bode plots were created for three PWM frequencies: 15, 25, and 40 Hz. A second-order transfer function, $\hat{G}(s)$, is used to estimate the frequency response. The bandwidth frequency, $\omega_{BW}$, is labeled for each data set.

## V. IMPEDANCE CONTROL

This work characterizes the bladders as an *impedance-style* device, as in some displacement, $x$, is measured, which is then converted to a desired output force, $F_{des}$. The force can be measured directly with a force sensor or indirectly with different sensors and a force model.

The sliding-mode tracking controller is designed to minimize tracking error in the presence of modeling uncertainty, which is a common approach in controlling pneumatic cylinders [21, 22, 23, 24, 25]. The error is defined as

$$e = F - F_{des} \quad (9)$$

where $F_{des}$ is the desired force trajectory.

A block diagram of the controller is shown in Fig. 10. The AFS robot is used to measure the bladder compression, $z$, and bladder force, $F$. The AFS is used to compress the bladder 20 mm at various compression rates. The compression value, $z$, is used to determine the desired force, $F_{des}$, based on a look up table that defines a desired haptic trajectory. A low-pass filter with a with a frequency of $\tau = 40$ Hz is used to minimize noise in the force measurement. This filter at this frequency is justified because the closed-loop bandwidth is below this value, meaning the actuator would not be capable of achieving the high frequency changes in force anyways. A dead zone is also implemented to only activate the controller should the error be sufficiently large. The dead zone removes unnecessary valve actuation when the error is sufficiently small. The sliding-mode controller takes error, $e$, and computes a desired flow, $Q_{des}$, which is then converted to $DC_{out}$ or $DC_{in}$ using (6) or (7), respectively. The drive circuitry then outputs the necessary PWM power signals to the solenoid valves.

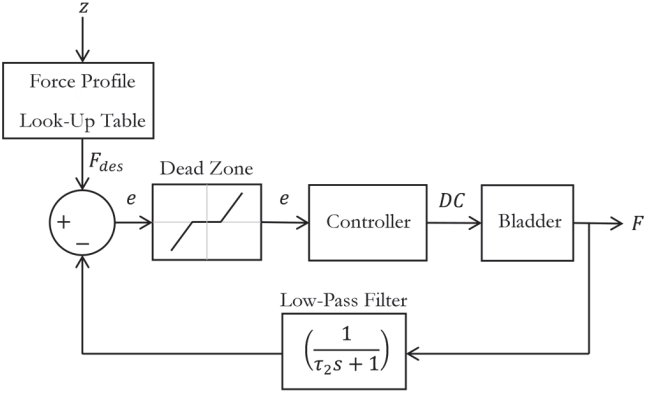

Fig. 10. The direct force controller takes actuator compression measurement, $z$, from the AFS and uses a lookup table to find the desired force, $F_{des}$. The controller, either PD or SMC, is used to minimize error $e$, which is the difference between the force measurement, $F$, and $F_{des}$. A low-pass filter and dead zone limit unnecessary valve actuation.

The derivation of the continuous sliding-mode tracking controller closely follows [26]. The remainder of the controller derivation follows the "off-the-shelf" procedure applied to the system states shown in (3). The control input, $Q$, for the SMC is written as

$$Q = -[\rho(z,\dot{z},P) + \beta_0]\tanh\left(\frac{e}{\mu}\right) \quad (10)$$

where $\beta_0$ and $\mu$ are gains. The term $\rho(z,\dot{z},P)$ is solved as

$$\rho(z,\dot{z},P) = \frac{a_{\max}(z,\dot{z},P) - \dot{F}_{des}}{b_{\min}(z,\dot{z},P)} \quad (11)$$

where $\dot{F}_{des}$ is the time derivative of the desired force, with terms

$$a_{\max}(z,\dot{z},P) = z\frac{\partial}{\partial z}\hat{f}_u + A_{b,u}|\dot{z}|\frac{P}{V_{\min}}\frac{\partial}{\partial P}\hat{f}_u \quad (12)$$

and

$$b_{\min} = \frac{P}{V_{\max}}\frac{\partial}{\partial P}\hat{f}_l. \quad (13)$$

Terms $\hat{f}_u$ and $\hat{f}_l$ are the upper and lower force bounds of $\hat{f}$, respectively, $V_{\min}$ and $V_{\max}$ are the minimum and maximum estimates of bladder volume, respectively, and $A_{b,u}$ is the upper estimate of the bladder top area in contact with the AFS. The specifically chosen upper and lower bounds of the uncertain terms ensure that the controller will always push the tracking error to zero. Additionally, the $\tanh()$ function is used in place of the traditional $\text{sgn}()$ function due to better performance from the $\tanh()$ function, which was found in [27].

## VI. EVALUATION

This section evaluates the ability of the bladder actuator to track various trajectories through the force vs. displacement workspace. The four proposed curves, $C_i$, are shown in Fig. 11.

Each curve is motivated by relevant haptic terrain display interactions. The first curve, $C_1$, simulates stepping on a soft shoe sole onto a hard surface such as concrete, which is the lower bound on $k_{\max}$, or the actuator stiffness required for successful haptic terrain display. Experimental testing of EVA foam, which is common in Nike running products, was done using the AFS robot. The linear stiffness was found to be approximated as 80 N/mm. The stiffness of soft soled shoes varies in the literature, ranging from 55 N/mm [28] to 257 N/mm [29]. A stiffness of 28.1 N/mm was used in $C_1$. Note that in an actual Smart Shoe design, four of these actuators could be combined in parallel under the heel portion of the shoe, bringing the total stiffness to 112.5 N/mm during heel contact, which is similar in value to the experimentally determined stiffness as well as the values published in the literature.

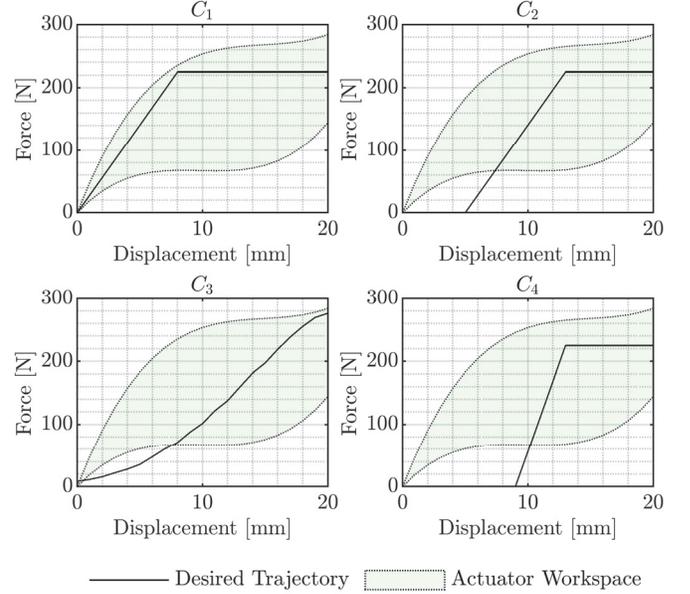

Fig. 11. The desired force vs. displacement curves: ($C_1$) Stiffness of a soft shoe sole, ($C_2$) stiffness of a soft shoe sole at a 5 mm displacement, ($C_3$) stiffness of dress shoe in sand, ($C_4$) double the stiffness of $C_1$ and $C_2$, but at 9 mm displacement. The bladder haptic workspace, according to the open-loop model, is shown in green.

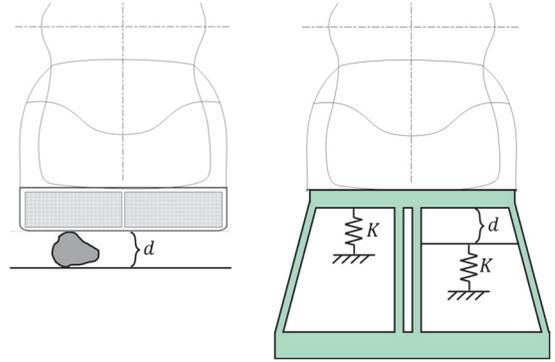

Fig. 9. (Left) An example of stepping on a stone in a normal shoe and (Right) stepping on the stone in a Smart Shoe. Two left bladder actuator simulates the stiffness of the shoe sole and stone, while the right actuator should display zero forces for a distance, $d$, before simulating the stiffness of the shoe sole.

The second curve, $C_2$, evaluates the same stiffness as $C_1$, but beginning at an offset of 5 mm. This example is motivated by Fig. 9. Simulating a user stepping on a small stone would require immediate force output from the actuator in the location of the stone, but the remaining actuators should output little to no force until the simulated shoe contacts the ground. Curve 3, $C_3$, shows experimentally collected data from a shoe being compressed into sand using the AFS robot. Finally, $C_4$, doubles the stiffness seen in $C_1$ and $C_2$, but at a greater offset of 9 mm. This example is chosen because the maximum stiffness of the actuator might not start at zero displacement, but further in the displacement workspace.

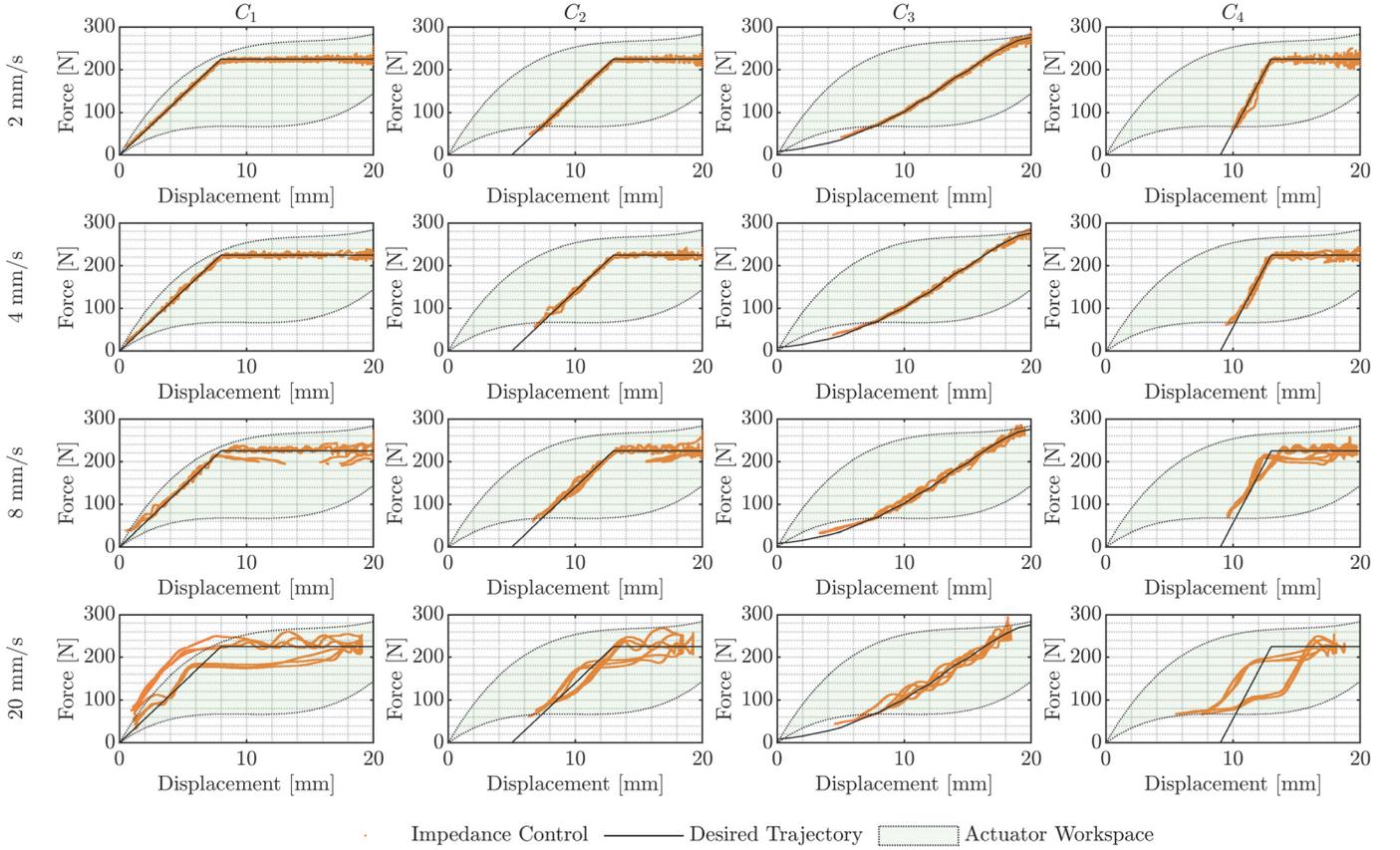

Fig. 12. The results of the impedance controller on the four workspace trajectories, $C_1$, $C_2$, $C_3$, and $C_4$. Light green shows the haptic workspace, black is the desired trajectory, and orange is the measured force and displacement.

To further evaluate the actuator in the domain of bandwidth, these curves are tested at four different compression rates: 2 mm/s, 4 mm/s, 8 mm/s, and 20 mm/s. The 20 mm/s compression rate compresses the entire bladder in one second, which is similar to what would be experienced during gait.

The set up in Fig. 3 is used for this evaluation. Sliding-mode controller gains of $(\beta_0, \mu) = (1 \times 10^{-5}, 5)$ were used for both input and output flow cases for $C_1$, $C_2$, and $C_3$, at the 2 mm/s, 4 mm/s and 8 mm/s compression rates. These were tuned to achieve a critically-damped response to a step input. The gain $\beta_0$ was increased to $\beta_0 = 3 \times 10^{-5}$ for $C_4$ because the doubled stiffness in $C_4$ essentially doubles the dynamics and $\beta_0$ must be increased to account for unmodeled dynamics. Additionally, for all 20 mm/s compressions, $\beta_0$ was increased to $5 \times 10^{-5}$, again, to compensate for unmodeled dynamics in the controller.

## VII. Results

The results of the evaluation are shown in Fig. 12 where the four different curves, $C_1$, $C_2$, $C_3$, and $C_4$, are plotted at the four compression rates, 2 mm/s, 4 mm/s, 8 mm/s, and 20 mm/s.

Table 1. The mean ($\mu$) and standard deviation ($\sigma$) of tracking error during direct force control.

| | $C_1$ | $C_2$ | $C_3$ | $C_4$ |
|---|---|---|---|---|
| $v$ [mm/s] | $\mu_e$ ($\sigma_e$) [N] | $\mu_e$ ($\sigma_e$) [N] | $\mu_e$ ($\sigma_e$) [N] | $\mu_e$ ($\sigma_e$) [N] |
| 2 | 2.5 (2.1) | 2.7 (2.3) | 2.1 (2.0) | 4.8 (4.4) |
| 4 | 3.2 (2.3) | 3.0 (2.8) | 2.2 (1.8) | 5.4 (4.1) |
| 8 | 8.0 (7.5) | 6.8 (5.8) | 3.6 (3.3) | 10.0 (8.9) |
| 20 | 22.3 (21.0) | 14.4 (10.2) | 8.6 (6.4) | 35.5 (34.1) |

Each plot includes the bladder workspace in light green with the control results shown in orange. Note that the measured control data is only included if the desired force and displacement reside within the bladder workspace. The mean and standard deviation of the tracking error, $\mu_e$ and $\sigma_e$, respectively, are shown in Table 1. Finally, two time response plots are included. First, Fig. 13 shows the desired flow, $Q_{des}$, computed by the sliding-mode tracking controller during the tracking of $C_3$ at 2 mm/s. The desired and measured forces are included. Second, Fig. 14 shows the same desired flow, desired force, and measured force, but for the tracking of $C_3$ at 20 mm/s.

## VIII. Discussion

The impedance control results in Fig. 12 and Table 1 show very small error for $C_1$, $C_2$, and $C_3$ at 2 mm/s, with a mean tracking error of less than 3 N, which is 1.2% of the 250 N workspace of the bladder actuator. The performance at 2 mm/s on $C_4$ has greater error, which was expected because the stiffness is double that of $C_1$ and $C_2$, which excites the unmodeled dynamics in the controller and lowers controller performance. At the rate of 4 mm/s, the tracking error remains under a mean of 3.5 N except for $C_4$ with a mean tracking error of 5.4 N. Tracking error increases at the 8 mm/s compression rate. At this point, the controller performance begins to decline on all four curves as the unmodeled dynamics become more relevant. Tracking $C_3$ gives the best results because at a constant compression rate, the change in the reference signal, $F_{des}$, remains low compared to the other curves with greater stiffness. Finally, the results for the 20 mm/s compression rate are tabulated. This compression rate was chosen to compress the bladder through its 20 mm workspace in 1 second, which is the correct order of magnitude of speeds expected during walking. The same general trend holds where $C_3$ results in the best performance due to $F_{des}$ experiencing the lowest dynamics. $C_4$ performs the worst, where the mean tracking error is up to 35.5 N, or 14.2% of the bladder workspace.

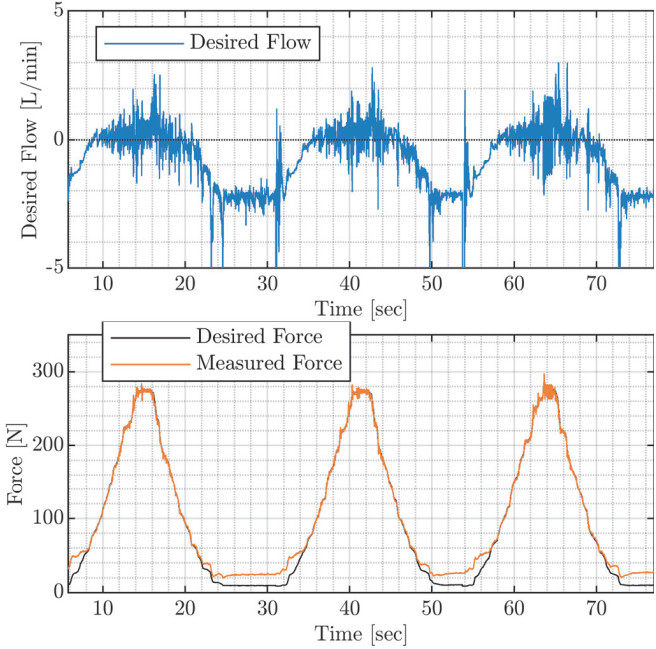

Fig. 13. The desired flow output from the controller as well as desired and measured force during the 2 mm/s compressions on curve $C_3$.

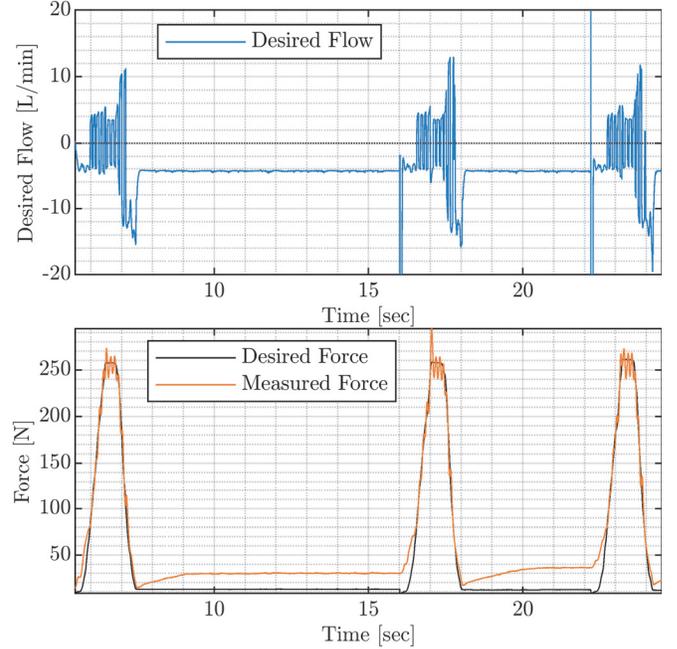

Fig. 14. The desired flow output from the controller as well as desired and measured force during the 20 mm/s compressions on curve $C_3$.

The 20 mm/s results also show some hysteresis effects that arise from a lack of actuator bandwidth. The results in Fig. 8 show the actuator closed-loop bandwidth reaches up to 7.3 Hz. Recall that actuator bandwidth should be compared to a value of 10 Hz, where most vertical ground reaction forces were observed. It is clear from this work that the current actuator design lacks some in bandwidth and should be improved in future versions. Bandwidth would be increased by increasing flow capability by either decreasing flow resistance or increasing pressure differential with a greater pressure supply or using a vacuum instead of atmospheric pressure for air flow output, which has been shown to have success in [30].

The heel strike transient discussed earlier in this work (reaching up to 75 Hz in [11]) must be addressed in future work. One possible solution would be to superimpose an open-loop transient over the control output to provide a transient response to the user [1]. A subject study would be required to determine if the open-loop transient is sufficiently close to a heel strike transient during gait.

Fig. 8 makes sense intuitively and further illustrates the tradeoffs in this work. With increasing PWM frequency, the control bandwidth increases, which will lead to better performance in tracking dynamic trajectories. However, this comes at the cost of potential increases in power consumption and heat while creating distracting audible and haptic sensations.

The results in Fig. 13 and Fig. 14 illustrate a few points of interest that occurred while tracking $C_3$. First is that even at the 20 mm/s, the controller generally tracks well, which is also true according to Fig. 12 and Table 1. Second, in Fig. 14 between 10 and 15 seconds, the desired flow, $Q_{des}$, is approximately -4 L/min, meaning the actuator is trying to release additional air to decrease force down to zero. However, the mechanical stiffness of the bladder walls results in a force of approximately 20 N, which is a 20 N tracking error. This occurs because the beginning of $C_3$ leaves the actuator workspace (which can be seen in Fig. 11). In fact, this behavior occurred for $C_2$ and $C_4$ as well. This illustrates that future iterations of these actuators should take great care to examine designs that minimize wall stiffness forces required to push the lower bound of the workspace closer to zero.

IX. CONCLUSION

This paper covers the identification of a force model, input and output flow models, and closed-loop bandwidth and closed-loop PWM control of soft-robotic bladders meant for haptic terrain display. It was found that the workspace size was sufficient to successfully simulate terrain. A closed-loop bandwidth of 7.3 Hz was found to be slightly lower than the goal of 10 Hz, but time response plots of the controller at compression rates of 20 mm/s were found to be promising. The results in this paper indeed could be integrated into a future Smart Shoe capable of impedance style haptic terrain display.

Future work should focus on integrating these bladders into a wearable haptic device, which would allow user evaluations. Since this paper used the AFS to measure force and displacement, Smart Shoe embedded systems need to be improved to provide sufficient measurement of bladder displacement, air pressure, and forces. The force modelling and controller should be efficient enough for microcontroller implementation, but microcontroller implementation should be verified in future work. Future work could also consider on improving bladder bandwidth and maximizing bladder workspace.

X. REFERENCES

[1] Springer, Springer Handbook of Robotics, B. Siciliano and O. Khatib, Eds., Springer-Verlag Berlin Heidelberg, 2008.

[2] T. H. Massie and J. K. Salisbury, "The PHANTOM Haptic Interface: A Device for Probing Virtual Objects," in *AMSE Symp. Haptic Interfaces for Virtual Environment and Teleoperated Systems*, Chicago, 1994.

[3] Novint, 2021. [Online]. Available: www.novint.com. [Accessed 14 May 2021].

[4] H. Son, I. Hwang, T.-H. Yang, S. Choi, S.-Y. Kim and J. R. Kim, "RealWalk: Haptic Shoes Using Actuated MR Fluid for Walking in VR," in *World Haptics Conference*, Tokyo, 2019.

[5] S. Bar-Haim, N. Harries, Y. B. M. Hutzler and I. Dobrov, "Training to Walk Amid Uncertainty with Re-Step: Measurements and Changes with Perturbation Training for Hemiparesis and Cerebral Palsy," *Disability and Rehabilitation Assistive Technology,* vol. 8, no. 5, pp. 417-425, 2013.

[6] Y. Wang and M. A. Minor, "Design and Evaluation of a Soft Robotic Smart Shoe for Haptic Terrain Rendering," *IEEE/ASME Trans. on Mechatronics,* vol. 23, no. 6, pp. 2974-2979, 2018.

[7] Y. Wang, T. E. Truong, S. W. Chesebrough, P. Willemsen, K. B. Foreman, A. S. Merryweather, J. M. Hollerbach and M. A. Minor, "Augmented Virtual Reality Terrain Display with Smart Shoe Physical Rendering: A Pilot Study," *IEEE Trans. on Haptics,* vol. 14, no. 1, pp. 174-187, 2021.

[8] Y. Wang, C. Gregory and M. A. Minor, "Improving Mechanical Properties of Molded Silcone Rubber for Soft Robotics Through Fabric Compositing," *Soft Robotics,* vol. 5, no. 3, pp. 272-290, 2018.

[9] K. B. Freckleton and M. A. Minor, "Modeling and Characterization of a Potential Bladder Based Orthotic Device to Mititgate Shoe Slip," in *IEEE Int. Conf. on Robotics and Automation (ICRA)*, Brisbane, 2018.

[10] J. E. Smeathers, "Transient Vibrations Caused by Heel Strike," *Proc Instn Mech Engrs,* vol. 203, pp. 181-186, 1989.

[11] J. T. Blackburn, B. G. Pietrosimone, M. S. Harkey, B. A. Luc and D. N. Pamukoff, "Comparison of Three Methods for Identifying the Heelstrike Transient During Walking Gait," *Medical Engineering and Physics,* vol. 38, pp. 581-585, 2016.

[12] J. J. Miller, "Design, Fabrication, and Evaluation of an Ankle Foot Simulator," University of Utah, MS Thesis, Salt Lake City, UT, 2016.

[13] J. Miller, A. Merryweather and M. Minor, "Design and Simulation of an Ankle Foot Simulator," in *SM3C*, Salt Lake City, UT, 2016.

[14] "Electro-Fluidic Systems Handbook 9th Edition," [Online]. Available: www.theleeco.com. [Accessed September 2019].

[15] M. Taghizadeh, A. Ghaffari and F. Najafi, "Modeling and Identification of a Solenoid Valve for PWM Control Techniques," *C. R. Mecanique,* vol. 337, pp. 131-140, 2009.

[16] N. Ye, S. Scavarda, M. Betemps and A. Jutard, "Models of Pneumatic PWM Solenoid Valve for Engineering Applications," *Trans. ASME,* vol. 114, pp. 680 - 688, 1992.

[17] S. Hodgson, M. Tavakoli, M. T. Pham and A. Leleve, "Nonlinear Discontinuous Dynamics Averaging and PWM-Based Sliding Control of Solenoid-Valve Pneumatic Actuators," *Trans. on Mechatronics,* vol. 20, no. 2, pp. 876-888, 2014.

[18] M.-C. Shih and M.-A. Ma, "Position Control of a Pneumatic Cylinder Using Fuzzy PWM Control Method," *Mechatronics,* vol. 8, pp. 241-253, 1998.

[19] V. Jouppila, S. A. Gadsden and A. Ellman, "Modeling and Identification of a Pneumatic Muscle Actuator System Controlled by an On/Off Solenoid Valve," in *Int. Fluid Power Conf.*, Aachen, 2010.

[20] V. T. Jouppila, S. Gadsden, B. Andrew, G. M., A. U. H. Ellman and S. R., "Sliding Mode Control of a Pneumatic Muscle Actuator System with a PWM Strategy," *Int. Journal Fluid Power,* vol. 15, no. 1, pp. 19-31, 2014.

[21] A. K. Paul, J. Mishra and M. Radke, "Reduced Order Sliding Mode Control for Pneumatic Actuator," *Trans. Control Systems Technology,* vol. 2, no. 3, pp. 271-276, 1994.

[22] T. Nguyen, J. Leavitt, F. Jabbari and J. E. Bobrow, "Accurate Sliding-Mode Control of Pneumatic Systems Using Low-Cost Solenoid Valves," *Transactions on Mechatronics,* vol. 12, no. 2, pp. 216-219, 2007.

[23] S. Pandian, Y. Hayakawa, Y. Kanazawa, Y. Kamoyama and S. Kawamura, "Practical Design of a Sliding Mode Controller f or Pneumatic Actuators," *Dynamic Systems, Measurement and Control,* vol. 119, no. 4, 1997.

[24] B. W. Surgenor and N. D. Vaughan, "Continuous Sliding Mode Control of a Pneumatic Actuator," *Dynamic Systems, Measurement and Control,* vol. 119, no. 3, 1997.

[25] K. Xing, J. Huang, Y. Wang, J. Wu, Q. Xu and J. He, "Tracking Control of Pneumatic Artificial Muscle Actuators Based on Sliding Mode and Non-Linear Disturbance Observer," *Iet Control Theory and Applications,* vol. 4, pp. 2058-2070, 2010.

[26] H. K. Khalil, Nonlinear Control, Upper Saddle River, NJ: Prentice Hall, 2015.

[27] G. Song and R. Mukherjee, "A Comparative Study of Conventional Nonsmooth Time-Invariant and Smooth Time-Varying Robust Compensators," *IEEE Transactions on Control Systems Technology,* vol. 6, no. 4, pp. 571-576, 1998.

[28] I. J. Wallace, E. Koch, N. B. Holowka and D. E. Lieberman, "Heel Impact Forces During Barefoot Versus Minimally Shod Walking Among Tarahumara Subsistence Farmers and Urban Americans," The Royal Society Publishing, 2018.

[29] M. Bishop, P. Fiolkowski, B. Conrad, D. Brunt and M. Horodyski, "Athletic Footwear, Leg Stiffness, and Running Kinematics," *Journal of Athletic Training,* 2006.

[30] A. A. Stanley, J. C. Gwilliam and A. M. Okamura, "Haptic Jamming: A Formable Geometry, Variable Stiffness Tactile Display Using Pneumatics and Particle Jamming," in *World Haptics Conference*, Daejeon, 2013.